# Automated PolyU Palmprint sample Registration and Coarse Classification

**Dhananjay D. M.[1], Dr C.V.Guru Rao[2] and Dr I.V.Muralikrishna[3]**

[1] Computer Science Department ,JNTU
Hyderabad, Andhra Pradesh, India

[2] Department of Computer Science &Engineering, SR Engineering College,
Warangal, Andhra Pradesh, India

[3] Former director R&D, JNTU
Hyderabad, Andhra Pradesh, India

**Abstract**
Biometric based authentication for secured access to resources has gained importance, due to their reliable, invariant and discriminating features. Palmprint is one such biometric entity. Prior to classification and identification registering a sample palmprint is an important activity. In this paper we propose a computationally effective method for automated registration of samples from PlolyU palmprint database. In our approach we preprocess the sample and trace the border to find the nearest point from center of sample. Angle between vector representing the nearest point and vector passing through the center is used for automated palm sample registration. The angle of inclination between start and end point of heart line and life line is used for basic classification of palmprint samples in left class and right class.
*Keywords:* Average filter, Binarization , Gaussian smoothing, Boundary tracking, Angle between vector, Gradient, Left palm print, Right palmprint.

## 1. Introduction

Biometric based personnel authentication has established as robust, reliable methodology. An automated biometric system is based on using invariant physiological or behavioral human characteristics for secured access [3]. Biometric trait such as fingerprint, signature, palmprint, iris, hand, voice or face can be used to authenticate a person's claim. Palmprint is one such biometric trait found to poses stable and unique discriminating features. A sample palmprint has many features such as, principle lines, datum point, ridges, delta point and minutiae features [7]. Due to lack availability of standardized palmprint capturing devices most of the research proposals are using PolyU palmprint database as baseline database, to compare and establish test results. The Biometric Research Centre (UGC/CRC) at The Hong Kong Polytechnic University has developed a real time palmprint capture device, and has used it to construct a large-scale palmprint database. The PolyU Palmprint Database contains 7752 grayscale images [9]. All the captured samples of PolyU database are aligned in a specific direction. Providing a computationally efficient method for palm print sample registration and coarse classification of sample palm print in order to reduce computation burden has motivated our paper. In this paper we propose a method for registration of PolyU palmprint database samples and classification into two basic classes. In all the discussion followed, palmprint sample refers to PolyU palmprint database sample.

This paper is organized in five sections. Section 1 is used for introduction. Key features of palmprint and preprocessing of sample palmprint is discussed in section 2. Method to establish boundary of palmprint along with sample center is presented in section 3. The process of finding out the required angle of rotation for palmprint sample alignment and registration is aimed at in section 4. Section 5 is used for finding the angle of inclination of heart line and life line for basic classification, followed by section 6 containing result and discussion.

## 2. Key features and pre processing of palmprint

### 2.1 Key features

Principle lines and datum points are regarded as useful palmprint key features and have been used successfully for verification. In addition, other features associated with a palmprint are palmprint geometrical features, wrinkle features, delta point features, ridges and minutiae features. All palmprints contains three prominent lines known as heart line, head line and life line. They are regarded as principle lines of the palm print. Heart line starts below little finger and ends below the index finger, Head line starts near the thumb region and ends below the heart line origination point. Lifeline encloses the thumb and adjoining region. Region I is an area enclosed by heart line, region II is an area enclosed by lifeline and region III is an area present below the heart line and enclosed by headline. Many a palmprint also contains line originating near wrist and dividing the head line and marching towards heart line. This line is known as fortune line. Fig.-1 shows a sample of palmprint and key aspects associated with it.

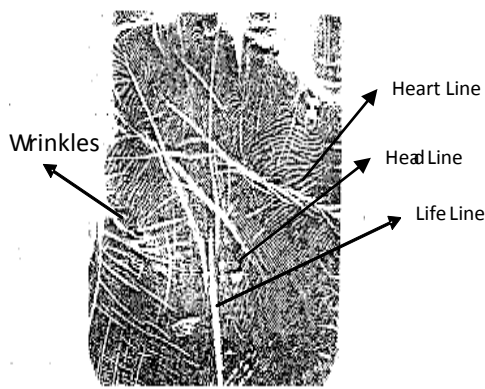

Fig1. Palmprint sample

## 2.2 Preprocessing of palm print sample

Submitted palmprint sample is submitted for preprocessing. This activity is used to smooth the given sample, obtain binarized sample and also to remove some noise present as additional objects in image. The result of this process is depicted in fig 2 (a,b). Steps involved in this process are as given below

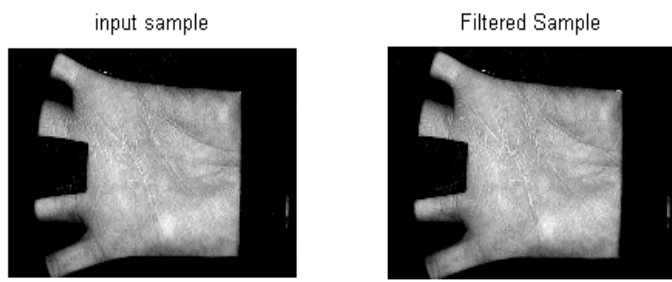

Fig 2a. Sample input and Average filtered sample

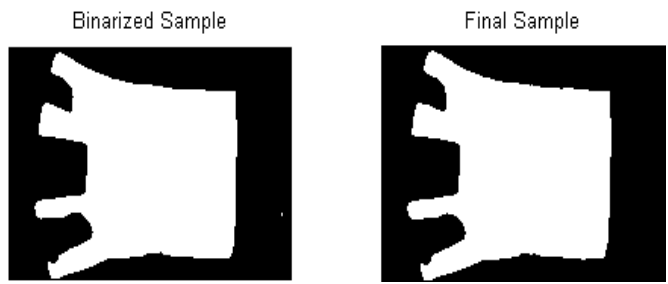

Fig 2b. Binarized image and Final Image

I. Apply 5x5 Gaussian low pass filter mask with standard deviation of value 0.10 to 0.25 on input image. The resulting image is smoothed image [6].

$$IMfilter(i, j) = IM(i, j) \otimes GaussianMask \quad (1)$$

*IM* (i,j) is input sample it is convolved with Gaussian mask to obtain filters image *IMfilter* (i,j) using (1).

II. Filtered palmprint sample obtained from (1) is submitted for binarization. We select the mean value of the filtered image as threshold and use (2) to convert into binary image.

$$IMbin(i, j) = 1 \text{ if } IMfilter(i, j) > \mu$$
$$IMbin(i, j) = 0 \text{ if } IMfilter(i, j) < \mu \quad (2)$$
$$\mu \text{ mean of filtered image}$$

III. Binarized sample images can contain objects which are of no interest and were retained after filtering. To remove such unwanted objects from image we apply labeling algorithm [8] and calculate area of each labeled objects. Object with largest area is retained as palmprint sample. This process is depicted in fig (2b). Following steps are used for this process.

(i) Obtain labeled image

IMLabel(i,j) = Label(IMbin(i,j)

(ii) Find number of objects with distinct label

Num=(IMlabel==x)

(iii) Calculate area of each object and retain object with maximum area.

(iv) Output image is logical '&'(and) of labeled image

## 3. Tracing boundary and establishing nearest point from center

3.1 Tracing Boundary of palmprint sample.

Output obtained after pre processing the image is used establish the boundary of the image by using suitable boundary tracing algorithm. In this paper we trace the boundary image by first establishing image coordinate with transition from 0 to 1 where 0 represents background and 1 represents object of interest. The neighborhood operation is used to trace the boundary of the image [8]. All co-ordinates representing boundary are collected in a boundary vector using (4). The boundary traced is being represented in fig3.

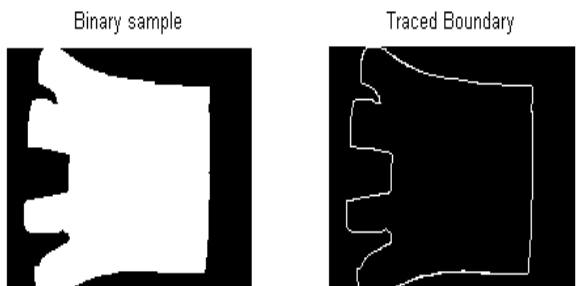

Fig3. Binary sample and traced boundary

$$VectB(i,k) = IMbin(i,j) \ if \ i,j \in boundary \quad (4)$$
$$k = 1,2$$

Further we establish center of the sample by finding of the center of mass of binary image using (5) & (6).

$$X0 = \sum VectB(i,1) / sizeofxcordinate \quad (5)$$
$$Y0 = \sum VectB(i,2) / sizeofxcordinate \quad (6)$$

3.2 Establishing nearest point from center

We use Euclidian distance measure to calculate the distance of points in vector VectB representing sample boundary using (7).

$$dist(i) = EuclidianDist([X0 Y0], VectB) \quad (7)$$

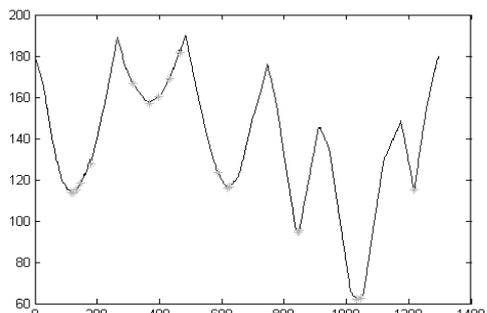

Fig4. Plot representing Euclidian distance from center

A plot of distance from center to bordering elements stored in vector VectB is represented in fig 4. The curve in graph represents curvaceous points of the given sample image. Using curve as an input we locate four points which are at minimum distance from center. Result of this operation is represented in fig 5.

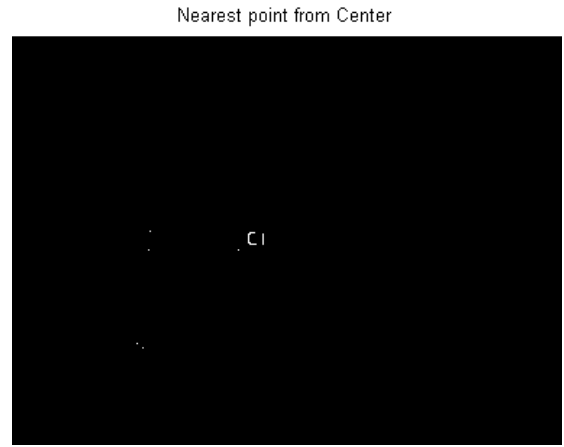

Fig5. Nearest points from center C1

## 4. Automated palm print alignment for registration

Given a vector V1 representing end of a line co-ordinates in Cartesian system and V2 another vector intersecting at point 'C1'. The angle between two vector intersecting at the given point is calculated by arctan of cross product of vector by dot product of vector [5]. We use (8c) to calculate the required angle.

$$V1 \times V2 = \|V1\|\|V2\|Sin\theta \quad (8a)$$

$$V1 \bullet V2 = \|V1\|\|V2\|\cos\theta \quad (8b)$$

$$\theta = \tan^\varphi (V1 \times V2)/(V1 \bullet V2) \quad (8c)$$

After establishing the center of palmprint sample, we pass a straight line through C1, this forms vector V1. The four nearest points established through the process discussed in previous section, we find out the nearest point from C1, this will establish the second vector V2. The output of the process is depicted in fig 6.

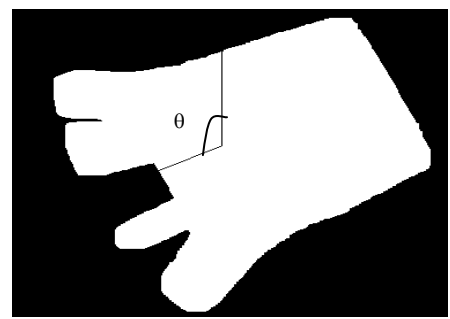

Fig6. Angle θ calculated for sample palm print

The input sample for which we calculated angle θ is rotated for uniform registration. This output is represented in fig7.

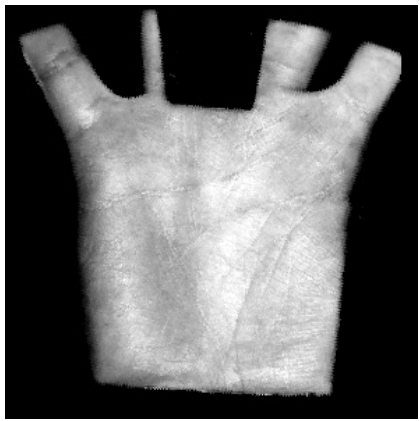

Fig7. Registered Palm print sample

## 5. Palm print sample course classification

### 5.1 Heart Line and Life line

Heart line, Head line and life line are considered as principle lines of palm print sample. Head line origination is from below the little finger and end near index finger. If a line is put from start point to end point its inclination can be observed in opposite direction in left a hand and right hand. Life line originates below the thumb region and encircles the thumb region and ends near the wrist. If a line is put from origination point to end points its inclination is in opposite direction for right and left hand. This property is shown in fig 8 using imaginary lines. We use this discriminating property of palm print sample to classify the sample in left & right palmprint sample class

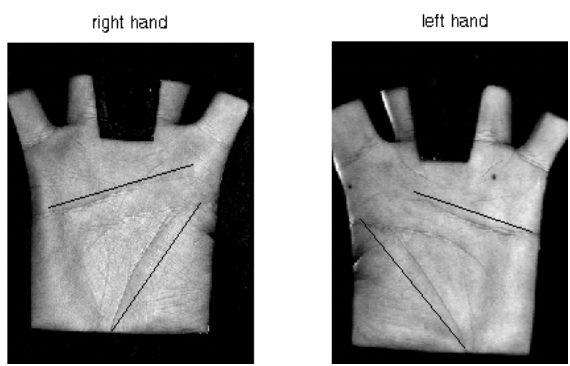

Fig8. Line inclination for left and right palmprint

### 5.2 Palm print sample classification

If P1 and P2 represent the origination and end point of heart line and Q1, Q2 represent the origination and end point of life line. We calculate the angle of inclination $\theta_1$, for line joining P1, P2. Let $\theta_2$ represent angle of inclination for points Q1, Q2. A value of $+\theta_1$ is present for right hand sample and $-\theta_1$ is present for left hand sample. Same holds true for life line using measured angle $\theta_2$.

Many of the palmprint sample may contain heart line which of shorter length and horizontal in nature. But all samples will contain life line prominently visible. We use two level checks to decide the class of the sample by using angle $\theta_1$ and $\theta_2$. If angle $\theta_1$ and $\theta_2$ are negative then class of sample is right class and if angle $\theta_1$ and $\theta_2$ is positive the class of sample is Left class. The complete process of sample registration with coarse classification is given in algorithm-1 as pseudo code

*Algorithm-1*

```
Sample=read_sample(palm database);
FilterSample=Apply(Gaussian mask on Sample);
M=mean(Sample);
BinSample=Sample>M;
Label(BinSample);
N=numofobject(BinSample);
Sample=N with Max area;
[x y]=center(sample)
VectB=Boundary(sample);
VectD=EcludianDist(x,y to VectB);
[p1 p2 p3 p4]=min(VectD)
Find Min of p1,p2,p3,p4
V1=[p1, [x y]]
V2=[[x y],0];
Theta=atan(V1*V2)/(V1.V2);
Sample=Rotate(Sample,Theta);
[P1 P2]=select(HeartLine);
[Q1 Q2]=select(Life Line);
Theta1=gradient(P1,P2);
Theta2=gradient(Q1,Q2)
If Theta1 & Theta2 <0
Class=Right
Else
Class=Left
Stop
```

## 6. Test result and conclusion

### 6.1 Test Results

Algorithm proposed here is implemented using Matlab7.0. Our primary source of palmprint sample database is PolyU database. Though all samples in the database are aligned in same direction, for testing we have applied image rotation by different angles to consolidate results. Fig9. (a, b, c, d, e, f, g) shows the results for subset of sample input

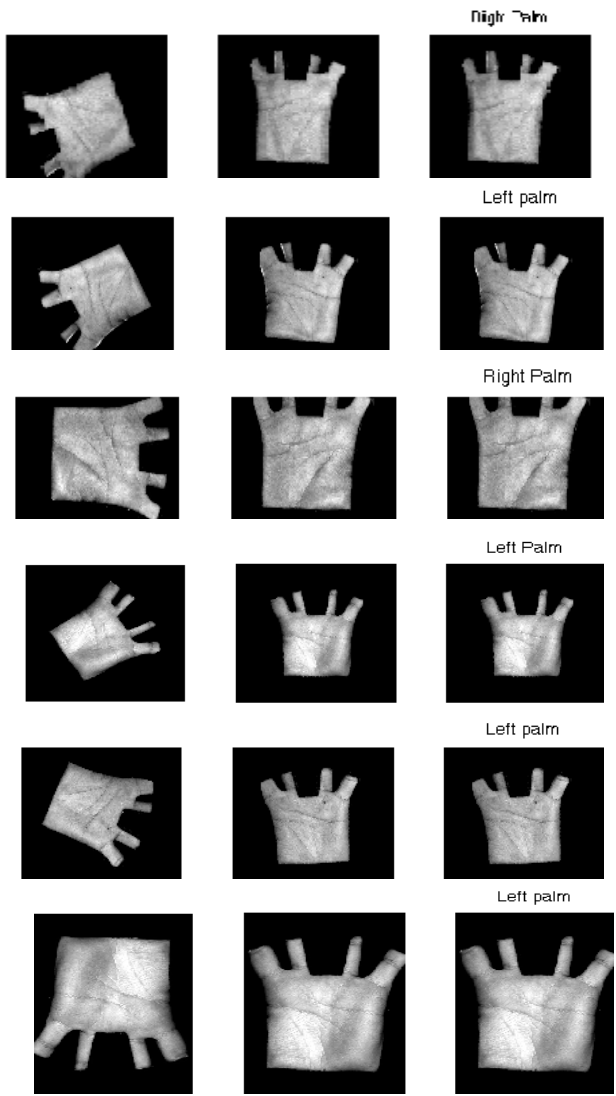

Fig9. (a,b,c,d,e,f) First column Sample input, Second column Aligned sample, Third column classified sample

## 6.2 Conclusion

Palm print samples archived in PolyU Palm print database are aligned in same direction as capturing device uses peg to restrict the movement. Our proposed algorithm can be used to allow user to obtain palm sample in any direction. The computational burden can be reduced, if primary classification is performed at acquisition level. The method proposed is invariant to transformation, which is useful feature to acquire palmprint sample from other devices.

**Acknowledgments**

We are thankful to "The Hong Kong Polytechnic University" for (PolyU) Palmprint Database (The Second Version) made available for research. Portions of the work tested on the PolyU Palmprint Database http://www.comp.polyu.edu.hk/~biometrics/

**Dhananjay D M** received his BE degree from GUG and has obtained M.Tech from VTU. Presently pursuing PhD under guidance of Dr C.V.Guru Rao and Dr I.V.Muralikrishna. He has presented paper in three conference and one journal. He is member of IETE,CSI India. His area of interest are Biometrics, Neural network, Pattern recognition.

**Dr. Guru Rao C V** He has distinguished himself as a teacher-researcher-administrator for more than 26 years in service of the student community and the society at large. Presently He is working as a Professor and Head, Department of Computer Science & Engineering at S R Engineering College, Ananthasagar, Warangal. He received a Bachelor's Degree in E&CEfrom Nagarjuna University, Guntur, India in the year 1981. He is double post graduate i.e., M.Tech in Electronic Instrumentation and M.E in Information Science & Engineering from Regional Engineering College, Warangal and  Motilal Nehru Regional Engineering College, Allahabad respectively. He was awarded a Ph.D Degree in CS&E IIT, Kharagpur, India in 2004. He had started his career at Kakatiya Institute of Technology & Science, Warangal, as a Lecturer in Electronics & Instrumentation Engineering in 1985. He had served the institute KITS Warangal in various capacities. He was elevated to the post of Principal of the same institute in 2007. He had served as a Member as well as  Chairperson, Board of Studies in Computer Science & Engineering and Information Technology, Kakatiya University, Warangal  and other institutions for several times. He is a Member on the Academic Advisory Committee of a Deemed University "Monad University, New Delhi" and Advisor of 21st Century Gurukulam, Kakatiya Unviersity, Warangal. He was nominated as a Member on to Industry Institute Interaction (III) Panel by A.P. State Council of Confederation of Indian Industry (CII), Andhra Pradesh. He was also nominated as a member of Engineering Agricultural Medicine Common Entrance Test (EAMCET) Admissions Committee and Post Graduate Engineering Common Entrance Test (PGECET) Committee in 2010 by Andhra Pradesh State Council of Higher Education, Hyderabad. He had published 53 technical research papers in various journals and conferences. A book titled "The Design and Analysis of Algorithms, 2e" by Anany Levitin was adapted by him in tune to the Indian standards and it was published by Pearson Education India. LAP LAMBERT Academic Publishing AG &

Co., Germany had proposed to publish his Ph.D. thesis "Design-For-Test Techniques for SOC Designs" in a book form. Under his guidance 11 students completed their Master's Dissertations including a scholar from Italy. Further, 25 research scholars are working on part-time basis to acquire a Ph.D. at different universities. He had been steering the "International Journal of Mathematics, Computer Science & Information Technology" in bringing out pragmatic research publications as Editor-in-Chief. Further, contributing to the "International Journal of Computational Intelligence Research and Applications" and "International Journal of Library Science" in publishing peer reviewed research articles as Member on the Editorial Board

**Dr I.V.Muralikrishna** M Tech(IIT-Madras) PhD (IISc-Bangalore) FIE, FIS, FAPASc, FICDM, MIEEE, FIGU. Has been on board of many prestigious institute. He was instrumental in implementation of many research project on weather modification, Environment and cloud seeding . He is receiver of many prestigious awards in recognition of his research work. He has been instrumental in organizing many international and national workshop , seminars, conferences. He was director of R & D at JNTU Hyderabad. Published 83 papers including 44 papers peer reviewed publications and Research Guidance to about 159 Ph D scholars and Graduate students. As on January 2011 Guided / Co-Guided 24 PhDs and 135 M Tech / MCA/ M Sc / MS in Faculty of Spatial Information Technology Faulty of Environmental Science and Technology Faculties of ECE Faculty of Civil Engg & Water Resources Faculty of Computer Science and Technology Faculty of Management Studies.